\documentclass[conference]{IEEEtran}
\IEEEoverridecommandlockouts
\usepackage{amsmath,amssymb,amsfonts}
\usepackage{algorithmic}
\usepackage{graphicx}
\usepackage{textcomp}
\usepackage{xcolor}

\usepackage{natbib}
\setcitestyle{numbers,square}

\usepackage{booktabs}
\usepackage{multirow}
\usepackage{enumitem}
\usepackage{tikz}
\usepackage{pythonhighlight}
\usepackage{listings}
\usepackage[utf8]{inputenc}
\usepackage{hyperref}

\def\BibTeX{{\rm B\kern-.05em{\sc i\kern-.025em b}\kern-.08em
    T\kern-.1667em\lower.7ex\hbox{E}\kern-.125emX}}
\begin{document}

\title{CLEAR-KGQA: Clarification-Enhanced Ambiguity Resolution for Knowledge Graph Question Answering}

\author{\IEEEauthorblockN{1\textsuperscript{st} Liqiang Wen}
\IEEEauthorblockA{\textit{Peking University} \\
Beijing, China \\
wenlq@pku.edu.cn }

\\

\IEEEauthorblockN{4\textsuperscript{th} Bing Li}
\IEEEauthorblockA{\textit{University of International} \\ \textit{Business and Economics} \\
Beijing, China \\
01630@uibe.edu.cn} 

\and

\IEEEauthorblockN{2\textsuperscript{nd} Guanming Xiong }
\IEEEauthorblockA{\textit{Peking University} \\
Beijing, China \\
gm\_xiong@pku.edu.cn}

\\

\IEEEauthorblockN{5\textsuperscript{th} Weiping Li}
\IEEEauthorblockA{\textit{Peking University} \\
Beijing, China \\
wpli@ss.pku.edu.cn}

\and

\IEEEauthorblockN{3\textsuperscript{rd} Tong Mo}
\IEEEauthorblockA{\textit{Peking University} \\
Beijing, China \\
motong@ss.pku.edu.cn}

\\

\IEEEauthorblockN{6\textsuperscript{th} Wen Zhao}
\IEEEauthorblockA{\textit{Peking University} \\
Beijing, China \\
zhaowen@pku.edu.cn}
}

\maketitle

\begin{abstract}
    This study addresses the challenge of ambiguity in knowledge graph question answering (KGQA).
    While recent KGQA systems have made significant progress, particularly with the integration of large language models (LLMs), they typically assume user queries are unambiguous, which is an assumption that rarely holds in real-world applications.
    To address these limitations, we propose a novel framework that dynamically handles both entity ambiguity (e.g., distinguishing between entities with similar names) and intent ambiguity (e.g., clarifying different interpretations of user queries) through interactive clarification.
    Our approach employs a Bayesian inference mechanism to quantify query ambiguity and guide LLMs in determining when and how to request clarification from users within a multi-turn dialogue framework.
    We further develop a two-agent interaction framework where an LLM-based user simulator enables iterative refinement of logical forms through simulated user feedback.
    Experimental results on the WebQSP and CWQ dataset demonstrate that our method significantly improves performance by effectively resolving semantic ambiguities.
    Additionally, we contribute a refined dataset of disambiguated queries, derived from interaction histories, to facilitate future research in this direction.\footnote{Code and data are available at: \url{https://github.com/JimXiongGM/CLEAR-KGQA}}
\end{abstract}

\begin{IEEEkeywords}
Knowledge Graph, Question Answering, Ambiguity
\end{IEEEkeywords}

\section{Introduction}

Knowledge graph question answering (KGQA) is a rapidly evolving research field that seeks to parse unstructured natural language (NL) into executable logical forms (e.g., SPARQL queries) to retrieve answers from a knowledge graph (KG). 

In recent years, numerous KGQA algorithms have been proposed to enhance the accuracy and efficiency of this process. 
For instance, studies such as \cite{Xiong-Guanming-ACL-2024-Interactive-KBQA}, \cite{Jiang-Jinhao-arXiv-2024-KG-Agent}, and \cite{Gu-Yu-arXiv-2024-Middleware} leverage the powerful reasoning capabilities of large language models (LLMs), such as GPT-4 \cite{OpenAI-2023-GPT4}, by treating these models as agents that interact with KGs through specialized tools in a multi-turn dialogue framework. 
Similarly, \cite{Haoran-Luo-arXiv-2023-ChatKBQA} demonstrates that fine-tuning open-source LLMs using extensive supervised data can also achieve remarkable performance. 
Beyond LLM-based approaches, earlier methods such as \cite{Das-Rajarshi-EMNLP-2021-Case-based-Reasoning} and \cite{Ye-Xi-ACL-2022-RNG-KBQA} focus on first retrieving subgraphs related to the query's topic entity and subsequently generating SPARQL queries.

However, most existing methods assume that user queries are well-formed and unambiguous, which rarely aligns with real-world scenarios. For instance, distinguishing between entities with similar names can pose significant challenges, as exemplified by the query ``What was Alice Walker famous for?'' where ``Alice Walker'' could refer to either a British fencer or an American author and activist. Ambiguities in user intent, such as the varied interpretations of ``famous for'' (e.g., awards received, notable works, or the individual’s profession), can further complicate the process. 
Such limitations are evident in the WebQSP dataset \cite{Yih-Wen-tau-ACL-2016-WebQSP}, which comprises queries collected from real users. State-of-the-art KGQA models achieve F1 scores in the range of 70–80 on WebQSP, but error analyses reveal that semantic ambiguity in user queries constrains further performance improvements.

To address these challenges, this study proposes several novel contributions:
\begin{itemize}[noitemsep]
    \item Building upon the tool-based design of \cite{Xiong-Guanming-ACL-2024-Interactive-KBQA}, we introduce an \texttt{AskForClarification} tool capable of requesting user clarification when needed.
    \item Employing Bayesian inference, we design a plug-and-play plugin to compute ambiguity scores during multi-turn interactions based on tool execution results.
    \item Developing an innovative two-agent interaction framework, where an LLM-based user simulator enables the QA agent to adaptively refine its logical forms in response to user feedback.
    \item Curating a refined dataset of disambiguated queries derived from the interaction history, which significantly enhances model performance. This dataset, characterized by clarified user intent, will be released to facilitate further research in the KGQA field.
\end{itemize}

\section{Related Work}

% KGQA
Knowledge graph question answering (KGQA) methods aim to parse natural language (NL) into executable logical forms (e.g., SPARQL queries) to retrieve answers from a knowledge graph (KG). Recent advances in KGQA leverage the powerful reasoning capabilities of large language models (LLMs) by treating them as agents that interact with KGs through carefully designed tools. These approaches achieve impressive results with minimal prompt engineering and few-shot learning.

% agent-env
Several works have explored different agent-environment interaction paradigms. Starting from topic entities, \cite{Jiang-EMNLP-2023-StructGPT} developed two specialized interfaces for accessing the KG, while \cite{Gu-Yu-arXiv-2024-Middleware} and \cite{Liu-Xiao-ICLR-2023-AgentBench} designed seven tools to facilitate this process. \cite{Jiashuo-Sun-ICLR-2024-Think-on-Graph} introduced a novel approach that enables LLMs to iteratively employ beam search reasoning on a KG.
\cite{Xiong-Guanming-ACL-2024-Interactive-KBQA} designed a set of tools based on graph patterns, enabling LLM agents to solve complex KGQA tasks through multi-turn interactions.
% other approaches
Furthermore, \cite{Zong-Chang-arXiv-2024-Triad} proposed assigning three distinct roles to the agent for addressing different subtasks, while \cite{Jiang-Jinhao-arXiv-2024-KG-Agent} constructed an instruction dataset based on existing KGQA datasets.
However, despite the impressive capabilities of LLMs, these methods have yet to fully address questions that require disambiguation and clarification.

% Ambiguity and Clarification in Dialogue Systems
Research on ambiguity and clarification in dialogue systems has been extensively studied. \cite{Keyvan-Huang-ACM-2022-Survey-Ambiguous-Queries} and \cite{Rahmani-Wang-ACL-2023-Survey-Asking-Clarification-Questions} provide comprehensive surveys on asking clarification questions in conversational systems.
% LLMs for Ambiguity Resolution
Recent works have explored leveraging LLMs to handle ambiguous queries. \cite{Liao-Yang-SIGIR-2023-Proactive-Conversational-Agents} introduces the concept of proactive conversational agents that can anticipate and address potential ambiguities. Building on this, \cite{Deng-Yang-EMNLP-Findings-2023-Proactive-Dialogues} proposes a proactive chain-of-thought prompting scheme to enhance LLMs' ability to identify and resolve ambiguities proactively.
\cite{Zhang-Tong-ACL-2024-CLAMBER} presents a systematic taxonomy for categorizing query ambiguities in open-domain question answering and introduces CLAMBER, a benchmark for evaluating LLMs' capabilities in identifying and clarifying ambiguous queries.

% Ambiguity in KGQA
In the context of KGQA, \cite{Xu-Jingjing-EMNLP-2019-Asking-Clarification-Questions} introduces CLAQUA, a dataset for handling ambiguous queries that supports determining clarification needs, generating questions, and predicting answers. However, without direct KG interaction, accurately determining when clarification is needed remains challenging.
\cite{Liu-Lihui-WWW-2023-PReFNet} proposes a Bayesian inference framework for handling ambiguous queries by decomposing posterior probabilities into more tractable components. While innovative, this method assumes given golden entities and requires substantial training data.

\section{Approach}

\begin{figure*}[htbp]
    \centering
    \includegraphics[width=1\textwidth,page=1]{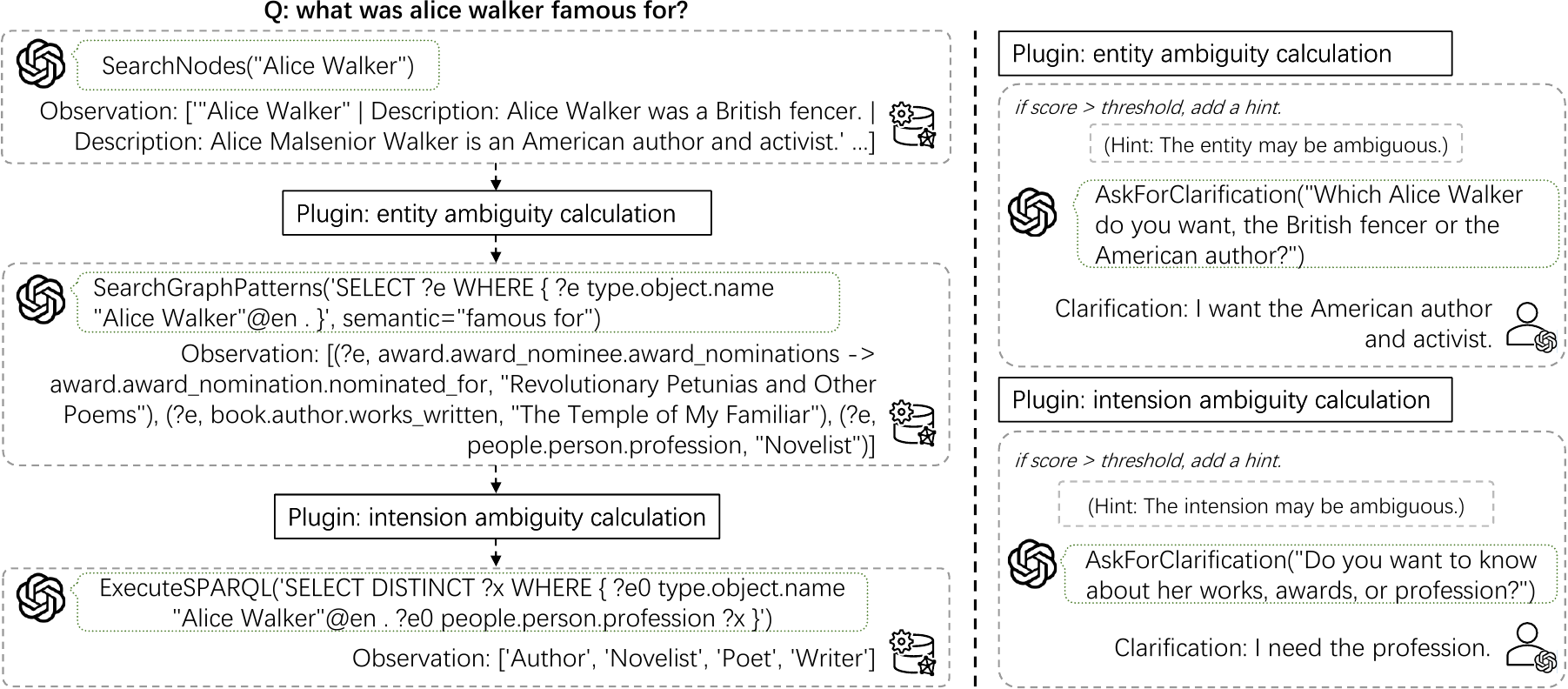}
    \caption{An example of the interactive process.}
    \label{fig:dialog_overview}
  \end{figure*}

\subsection{Problem Formulation}
This study addresses the challenge of ambiguity in knowledge graph question answering (KGQA) by quantifying and resolving uncertainties in both entity recognition and intent interpretation. Given a natural language question $Q$ and a knowledge graph $\mathcal{G}$, our goal is to generate an executable SPARQL query $S$ while explicitly handling potential ambiguities. Formally, we aim to model:

\begin{equation}
p(S|Q, \mathcal{G}, \mathcal{C}),
\end{equation}

where $\mathcal{C}$ represents the user clarification information obtained through interaction. The knowledge graph $\mathcal{G}$ consists of triples $(e_h, r, e_t) \in E \times R \times (E \cup L)$, where $E$ denotes the set of entities, $R$ signifies the set of relations, and $L$ includes literal values.

The key challenge lies in quantifying two types of ambiguity:

\begin{itemize}
    \item Entity ambiguity: Given a set of candidate entities $E_{\text{obs}} = \{e_1, ..., e_N\}$ obtained from the execution results of \texttt{SearchNodes}, we compute an ambiguity score $\text{Ambiguity}_E(Q, E_{\text{obs}})$ that measures the uncertainty in entity linking.
    
    \item Intent ambiguity: Given a set of predicted topic entities $E_q'$ and a set of candidate predicate-entity pairs $R_{\text{obs}} = \{(r_1, e_1'), ..., (r_N, e_N')\}$ obtained from the execution results of \texttt{SearchGraphPattern}, we calculate $\text{Ambiguity}_I(Q, E_q',R_{\text{obs}})$ to assess the uncertainty in predicate selection.
\end{itemize}

These ambiguity scores guide an interactive clarification process, where the agent may request additional information from user when the scores exceed predefined thresholds. The final SPARQL query $S$ is generated by incorporating both the original question and any clarification feedback.

\subsection{Overview}

Recent advancements in large language models (LLMs) have demonstrated remarkable capabilities in few-shot learning and reasoning. However, effectively handling ambiguity in KGQA remains a significant challenge, even with LLMs. To address this gap, we introduce clarification-enhanced ambiguity resolution (CLEAR-KBQA), an interactive framework that leverages LLMs to explicitly quantify and resolve ambiguities through multi-turn interaction. 

As illustrated in Figure \ref{fig:dialog_overview}, we introduce a novel \texttt{AskForClarification} tool that facilitates direct user interaction. 
However, relying solely on tool descriptions in prompts, LLM's tool invocation decisions remain uncontrollable in each interaction round. 
Therefore, we design a plugin that quantifies entity and intent ambiguity to guide LLM's tool usage.
The system uses a two-agent framework to collaboratively resolve ambiguities through natural dialogue.

\subsection{Tools Design}

Following \cite{Xiong-Guanming-ACL-2024-Interactive-KBQA}, we adopt three tools and introduce a new clarification tool:

\textbf{SearchNodes(name)}: This tool performs entity linking by searching for nodes in the knowledge base using a given surface name, returning both formal entity names and their distinguishing features such as descriptions and types.

\textbf{SearchGraphPattern(sparql, semantic)}: This tool identifies and ranks relevant graph predicates by querying one-hop subgraphs around specified entities, optimizing for semantic relevance and handling complex structures like CVT nodes.

\textbf{ExecuteSPARQL(sparql)}: This tool executes SPARQL queries directly against the knowledge base, providing fundamental query capabilities.

\textbf{AskForClarification(text)}: This newly introduced tool manages interactive disambiguation by generating natural language clarification requests when ambiguity is detected. The tool interrupts the interaction process and waits for user input before proceeding. Examples are provided in Figure \ref{fig:dialog_overview}.

\subsection{Interactive Process}

\paragraph{Question Answering Agent} is designed to generate the SPARQL query with multi-turn interaction with the KG. In each turn $T$, it generates an action based on the $\mathrm{Prompt_{qa}}$ and the history $H$ of the interaction. Specifically, this procedure follows:

\begin{equation}
    \label{eq:interactive_process_qa}
    a_T = \text{LLM}(\{\mathrm{Prompt_{qa}}, H\})
\end{equation}

\begin{equation}
    H = \{a_0,o_0,c_0, ...,a_{T-1}, o_{T-1}, c_{T-1}\}
\end{equation}

where $\mathrm{Prompt_{qa}} = \{\text{Inst}, E, Q\}$, consists of pre-written tool descriptions, tool usage, and the format of interaction. $E$ denotes a set of exemplars. $a$ is an action, and $o$ is the observation defined as $o_T = \text{Tool}(a_T)$, and $c$ denotes the clarification response from the user.

% Dummy User Agent: 给定SPARQL，对于每一个input clarification，生成一个clarification response C。
\paragraph{Dummy User Agent} is designed to simulate the user's response to the clarification request from the question answering agent. Given the golden SPARQL query $S'$ and a clarification request, the agent generates a clarification response $c_t$ at turn $t$. The formulation is as follows:

\begin{equation}
    \label{eq:interactive_process_dummy_user}
    c_t = \text{LLM}(\{\mathrm{Prompt_u}, S', a_t\})
\end{equation}

\subsection{Clarification Plugin}

Based on the interaction results from each turn, we design a plugin to quantify ambiguity. 
When the ambiguity score exceeds a predefined threshold, a hint is incorporated into the observation to explicitly guide the LLM to clarify.

\subsubsection{Entity Ambiguity}  
To assess the ambiguity of entities retrieved by the \texttt{SearchNodes} tool relative to a given question $Q$, a Bayesian-based metric is designed to determine whether the entities in $E_{\text{obs}} = \{ e_1, e_2, \dots, e_N \}$ are ambiguous. 
Our goal is to compute $P(e \mid Q)$.
First, the prior probability of each entity $e_i$ is computed using its popularity $u_{e_i}$ in a knowledge base (e.g., Freebase):

\begin{equation}
P(e_i) = \frac{u_{e_i}}{\sum_{j=1}^{N} u_{e_j}}
\end{equation}

The conditional probability $P(Q \mid e_i)$, which measures the relevance of $e_i$ to $Q$, is derived via the description $\text{Desc}(e_i)$ of $e_i$ using a LLM. Specifically:

\begin{equation}
P(Q \mid e_i) \propto \frac{1}{\text{PPL}(Q, \text{Desc}(e_i))}
\end{equation}

where $\text{PPL}(\cdot)$ represents the perplexity function. This is normalized using a softmax function to ensure a valid probability distribution. The posterior probability $P(e_i \mid Q)$ is then computed via Bayes' theorem:

\begin{equation}
P(e_i \mid Q) \propto P(Q \mid e_i) P(e_i)
\end{equation}

The entropy of this distribution, $H = -\sum_{i=1}^{N} \tilde P(e_i \mid Q) \log \tilde P(e_i \mid Q)$, where $\tilde P(e_i \mid Q)$ is the softmax normalization of $P(e_i \mid Q)$ over all entities, quantifies the uncertainty or ambiguity of the retrieved entities.
The ambiguity score is obtained by dividing $H$ by the maximum possible entropy, $\log N$:

\begin{equation}
\text{Ambiguity}_E(Q, E_{\text{obs}}) = \frac{H}{\log N}
\end{equation}

This score ranges from 0 to 1. A score close to 0 means low ambiguity (a single entity clearly matches question $Q$). A score close to 1 means high ambiguity (multiple entities are similarly relevant), requiring clarification.

\subsubsection{Intent Ambiguity}  
To evaluate the ambiguity of predicates retrieved using the \texttt{SearchGraphPattern} tool, a metric is employed to assess the uncertainty in the relationship between a predicted entity $e_q'$ and a question $Q$. 
Let $R_{\text{obs}} = \{ (r_1, e_1'), (r_2, e_2'), \dots, (r_N, e_N') \}$ represent the list of candidate predicate-tail entity pairs, where $r_i$ is a predicate and $e_i'$ is the corresponding tail entity sampled from the KG. 
The goal is to compute $P(r \mid Q)$, which is simplified via Bayes' theorem:

\begin{equation}
P(r_i \mid Q) \propto P(Q \mid e_q', r_i, e_i') P(e_q', r_i, e_i')
\end{equation}

Here, $P(Q \mid e_q', r_i, e_i')$ represents the likelihood of $Q$ given the predicate $r_i$, its tail entity $e_i'$, and the predicted entity $e_q'$. The prior $P(e_q', r_i, e_i')$ captures the joint significance of these components for interpreting $r_i$. Following the same perplexity-based approach as for entities, the ambiguity score for predicates is given by:

\begin{equation}
\text{Ambiguity}_I(Q, R_{\text{obs}}) = \frac{-\sum_{i=1}^{N} \tilde{P}(r_i) \log \tilde{P}(r_i)}{\log N}
\end{equation}

where $\tilde{P}(r_i)$ is the softmax normalization of $P(r_i \mid Q)$ over all predicates.
Similarity, a high score indicates that multiple predicates exhibit similar relevance, requiring further clarification.

\subsection{Unambiguous Dataset Construction}

We use the interaction history of the dummy user agent to construct unambiguous questions. To mitigate the ambiguity inherent in the original question, we leverage both the golden SPARQL query $S'$ and the interaction history $H_u = \{a_0,c_0, ..., a_{T-1}, c_{T-1}\}$ to construct a refined question. The formulation is:

\begin{equation}
    \label{eq:unambiguous_question_generation}
    Q_u = \text{LLM}(\{\mathrm{Prompt_{gen}}, S', H_u\})
\end{equation}

\section{Experiment}

% 4.1
\subsection{Dataset}

\begin{table}[ht]
    \caption{Dataset Statistics}
    \centering
    \scalebox{1}{

    \begin{tabular}{ccc}
        \toprule
        \textbf{Dataset} & \begin{tabular}[c]{@{}c@{}}\textbf{\#Instance}\\      \textbf{(Train/Test)}\end{tabular} & \begin{tabular}[c]{@{}c@{}}\textbf{Raw \#Instance}\\      \textbf{(Train/Dev/Test)}\end{tabular} \\ \midrule
        WebQSP & 100 / 300 & 3,098 / - / 1,639 \\
        CWQ & 200 / 600 & 27,639 / 3,519 / 3,531 \\ \bottomrule
    \end{tabular}

    }
\label{tab:dataset_stats}
\end{table}

\textbf{WebQuestionsSP} (WebQSP) \cite{Yih-Wen-tau-ACL-2016-WebQSP} is a widely used dataset in KGQA research. The questions were collected from the Google Suggest API, and the corresponding SPARQL queries were annotated by human based on Freebase. Since WebQSP was sampled from real-world scenarios, the questions exhibit significant ambiguity.

\textbf{ComplexWebQuestions 1.1} (CWQ) \cite{Talmor-Alon-NAACL-2018-ComplexWebQuestions-CWQ} extends WebQSP by sampling questions based on four types of complex SPARQL templates and having humans paraphrase them into natural language questions.

WebQSP questions can be classified into 1-hop and 2-hop categories based on the length of the inferential relation chain. CWQ questions are categorized into four types: Conjunction (Conj), Composition (Compo), Comparative (Compa), and Superlative (Super). 
Following \cite{Xiong-Guanming-ACL-2024-Interactive-KBQA}, we adopt a low-resource setting by using 50 annotated multi-turn samples per question category for training. Evaluation is also conducted on a sampled test dataset, ensuring consistency with the prior work.
The dataset statistics are shown in Table \ref{tab:dataset_stats}.

\subsection{Implementation Details}

For the interactive process, we employ OpenAI's GPT-4o (\texttt{gpt-4o-2024-08-06}) as our LLM agent.
For calculating ambiguous scores and conducting fine-tuning experiments with open-source large language models (open-LLMs), we utilize \texttt{Llama-3.1-8B-Instruct} \cite{Aaron-Grattafiori-etal-2024-Llama-3-Herd-of-Models} as the base model. 
All experiments are conducted on two NVIDIA A100 80GB GPUs. The hyperparameters setting is followed \cite{Xiong-Guanming-ACL-2024-Interactive-KBQA}.

% 4.2
\subsection{Baselines}

\begin{table*}[htp]
    \caption{Results on WebQSP and CWQ.}
    \centering
    \scalebox{1}{

    \begin{tabular}{lcccccccccc}
    \toprule
    \multicolumn{1}{c}{\multirow{2}{*}{\textbf{Method}}} & \multicolumn{4}{c}{\textbf{WebQSP}} & \multicolumn{6}{c}{\textbf{CWQ}} \\ \cline{2-11} 
    \multicolumn{1}{c}{} & \textbf{1-hop} & \textbf{2-hop} & \textbf{Overall} & \textbf{RHits@1} & \textbf{Conj} & \textbf{Compo} & \textbf{Compa} & \textbf{Super} & \textbf{Overall} & \textbf{EM} \\ \midrule
    \multicolumn{11}{c}{Prior FT SOTA} \\ \midrule
    DeCAF (w/ full data) & 74.72 & 76.32 & 75.52 & 80.28 & 69.19 & 53.54 & 18.04 & 28.00 & 42.19 & 50.83 \\
    \quad w/ small data & 39.96 & 44.02 & 44.02 & 44.98 & 24.00 & 29.86 & 38.42 & 37.78 & 32.51 & 35.33 \\ \midrule
    \multicolumn{11}{c}{Prompting w/ GPT-4} \\ \midrule
    IO & 28.54 & 50.05 & 39.29 & 45.51 & 47.54 & 29.71 & 33.66 & 24.67 & 33.89 & 45.67 \\
    CoT & 26.66 & 51.35 & 39.01 & 47.08 & 50.65 & 28.78 & 36.98 & 29.78 & 36.55 & 47.50 \\ \midrule
    \multicolumn{11}{c}{Multi-turn Interaction w/ GPT-4} \\ \midrule
    Interactive-KBQA & 69.99 & 72.41 & 71.20 & 72.47 & 47.44 & 59.00 & 47.89 & 41.96 & 49.07 & 59.17 \\
    \quad w/ Golden Entity & 77.50 & 79.79 & 78.64 & 79.25 & 53.36 & 68.19 & 52.73 & 52.69 & 56.74 & 66.50 \\
    Ours & \textbf{84.50} & \textbf{81.51} & \textbf{83.00} & \textbf{83.00} & \textbf{43.30} & \textbf{64.73} & \textbf{50.78} & \textbf{51.01} & \textbf{52.46} & \textbf{66.33} \\ \midrule
    \multicolumn{11}{c}{Multi-turn Interaction w/ FT open-LLM} \\ \midrule
    Interactive-KBQA & 61.68 & 55.22 & 58.45 & 59.45 & 37.00 & 32.79 & 58.51 & 54.60 & 45.73 & 48.83 \\
    \quad w/ Golden Entity & 70.18 & 63.90 & 67.04 & 69.73 & 42.19 & 43.01 & 62.54 & 61.71 & 52.36 & 57.83 \\
    Ours & \textbf{76.22} & \textbf{71.31} & \textbf{73.77} & \textbf{73.77} & \textbf{45.00} & \textbf{46.96} & \textbf{64.93} & \textbf{66.53} & \textbf{55.86} & \textbf{62.33} \\ \midrule
    \multicolumn{11}{c}{Un-Amb Dataset} \\ \midrule
    Interactive-KBQA (GPT-4) & 86.93 & 79.43 & 83.18 & 83.18 & 46.54 & 66.64 & 60.52 & 47.01 & 55.19 & 70.52 \\
    \quad w/ open-LLM & 87.60 & 76.11 & 81.86 & 81.86 & 55.63 & 60.20 & 70.75 & 53.20 & 59.97 & 67.85 \\ \bottomrule
    \end{tabular}

}
\label{tab:main_res}
\end{table*}

To comprehensively evaluate our approach, we have selected a range of prior state-of-the-art (SOTA) baseline models.

\textbf{Prior Fine-tuning (FT) SOTA}. We selected DeCAF \cite{Yu-Donghan-ICLR-2023-DecAF}, a semantic parsing (SP)-based method, as our baseline. DeCAF leverages T5 \cite{Raffel-JMLR-2020-T5} to generate both logical forms and direct answers. For fair comparison, we re-trained the model on our small training dataset.

\textbf{Prompting based on GPT-4}. We employed direct input-output prompting (IO) and Chain-of-Thought prompting (CoT prompt) \cite{Wei-Jason-NeurIPS-2022-ChainOfThought} as our prompting baselines.

\textbf{Multi-turn Interaction method}. We selected Interactive-KBQA \cite{Xiong-Guanming-ACL-2024-Interactive-KBQA}, a GPT-4-based multi-turn interaction approach. This method leverages LLMs as agents that iteratively interact with the KG to perform semantic parsing and generate SPARQL queries. Furthermore, we fine-tuned open-LLMs as alternatives to GPT-4.

% 4.3
\subsection{Evaluation Metrics}

We use average F1 score as the primary metric. We also report Random Hits@1 (RHits@1) \cite{Shu-Yiheng-EMNLP-2022-TIARA} and Exact Match (EM) \cite{Talmor-Alon-NAACL-2018-ComplexWebQuestions-CWQ} for reference.

% 4.4
\subsection{Results}

Table \ref{tab:main_res} presents comprehensive comparisons.
Compared to the GPT-4-based baseline, our method achieves 83.00 and 52.46 on WebQSP and CWQ respectively, representing improvements of 11.8 and 3.39 percentage points.
Moreover, when using fine-tuned open-LLM as the baseline, our method demonstrates even larger improvements of 15.32 and 10.13 percentage points on WebQSP and CWQ respectively.
Since the WebQSP dataset contains real user queries, our model shows substantial improvements after intent clarification. While CWQ focuses on complex questions, we were limited to using 1-shot prompting to instruct the LLM about the interaction format due to context window constraints. Although this makes it challenging to cover all cases, our plugin still helps improve performance by 3.39 points, demonstrating its effectiveness.

We also conducted experiments using golden entities for the baseline to isolate the impact of entity ambiguity. As mentioned above, since CWQ questions are template-generated complex queries, they exhibit less ambiguity. Our method introduces a clarification action that expands the model's action space. Consequently, in the prompting-based approach, the baseline with golden entities performs better. However, in the fine-tuned model setting for CWQ, where the model has learned from hundreds of samples and developed more robust question pattern recognition, adding our plugin outperforms the golden entity baseline by 3.5 percentage points, further validating the effectiveness of our approach.

\subsection{Analysis of the Unambiguous Dataset}

\begin{table}[ht]
    \caption{Statistics of the unambiguous dataset.}
    \centering
    \scalebox{1}{

    \begin{tabular}{ccccc}
    \toprule
    \textbf{Dataset} & \textbf{Ave. \#Entity} & \textbf{Ave. \#Intent} & \textbf{\#item} & \textbf{Persent} \\ \midrule
    WebQSP & 0.47 & 0.67 & 234 & 78.00 \\
    CWQ & 0.14 & 0.48 & 331 & 55.17 \\ \bottomrule
    \end{tabular}

    }
\label{tab:stats_unambiguous_dataset}
\end{table}

\begin{table}[ht]
    \caption{Results on the unambiguous dataset.}
    \centering
    \scalebox{1}{

    \begin{tabular}{lcccc}
        \toprule
        \multicolumn{1}{c}{\multirow{2}{*}{\textbf{Dataset}}} & \multicolumn{2}{c}{\textbf{WebQSP}} & \multicolumn{2}{c}{\textbf{CWQ}} \\ \cline{2-5} 
        \multicolumn{1}{c}{} & \textbf{Overall} & \textbf{RHits@1} & \textbf{Overall} & \textbf{EM} \\ \midrule
        \multicolumn{5}{c}{GPT-4} \\ \midrule
        Origin & 71.20 & 72.47 & 49.07 & 59.17 \\
        New dataset & 83.18 & 83.18 & 55.19 & 70.52 \\
        Gain & 11.98 & 10.71 & 6.12 & 11.35 \\ \midrule
        \multicolumn{5}{c}{FT open-LLM} \\ \midrule
        Origin & 58.45 & 59.45 & 45.73 & 48.83 \\
        New dataset & 81.86 & 81.86 & 59.97 & 67.85 \\
        Gain & 23.41 & 22.41 & 14.24 & 19.02 \\ \bottomrule
    \end{tabular}

    }
\label{tab:res_unambiguous_dataset}
\end{table}

% Statistics
Using multi-turn interaction histories based on GPT-4, we generated unambiguous questions from the WebQSP and CWQ test datasets, resulting in a unambiguous version (Un-Amb). The dataset statistics are shown in Table \ref{tab:stats_unambiguous_dataset}.

In the WebQSP dataset, we observe an average of 0.47 entity ambiguities and 0.67 intent ambiguities per item, with 78 percent of questions requiring regeneration. In contrast, only 55.17 percent of questions in the CWQ dataset needed regeneration. This difference indirectly validates that the WebQSP dataset exhibits more significant ambiguity phenomena.
% Performance Improvement 
We evaluated the baseline Interactive-KBQA method on the Un-Amb, with results shown in Table \ref{tab:res_unambiguous_dataset}. The results demonstrate substantial performance improvements even without our plugin. Notably, in the WebQSP dataset using the FT open-LLM setting, we observed a remarkable 23.41 percentage point improvement in performance.

\subsection{Impact of Threshold Selection}

\begin{figure}[ht]
    \centering
    \includegraphics[width=\linewidth,height=0.4\textheight,keepaspectratio]{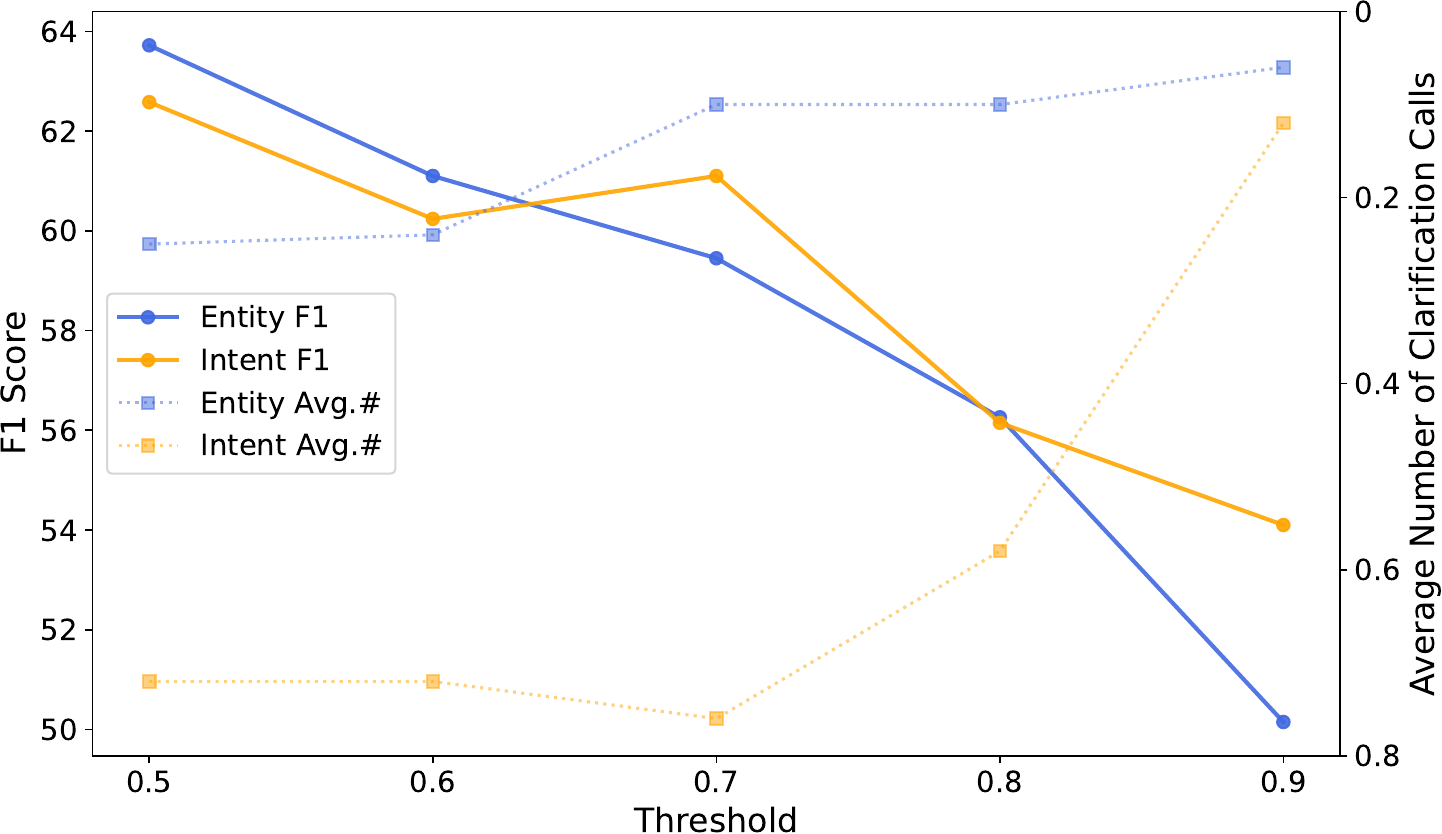}
    \caption{Grid search results on the threshold selection.}
    \label{fig:threshold_analysis}
\end{figure}

The thresholds for entity score and intent score are crucial parameters, as they determine when the plugin explicitly hints to the LLM to initiate clarification, thereby influencing the interaction process. To identify optimal thresholds, we conducted a grid search experiment by varying the thresholds from 0.5 to 0.9 (while fixing the other parameter at the default value of 0.5). Specifically, we merged the WebQSP and CWQ datasets and randomly sampled 100 instances (WQCWQ) as a validation set.

The experimental results are shown in Figure \ref{fig:threshold_analysis}. The left y-axis represents the F1 score, demonstrating that lower thresholds lead to more frequent clarification hints from the plugin, resulting in higher accuracy. The right y-axis shows the average number of clarification rounds per data sample (inverted scale), indicating that lower thresholds correspond to increased user interactions.
An effective system should minimize user interactions while maintaining high accuracy. Based on these considerations, we selected an entity score threshold of 0.6 and an intent score threshold of 0.8 for both datasets.

\subsection{Distribution of Ambiguity Score}

\begin{figure}[ht]
    \centering
    \includegraphics[width=\linewidth,height=0.5\textheight,keepaspectratio]{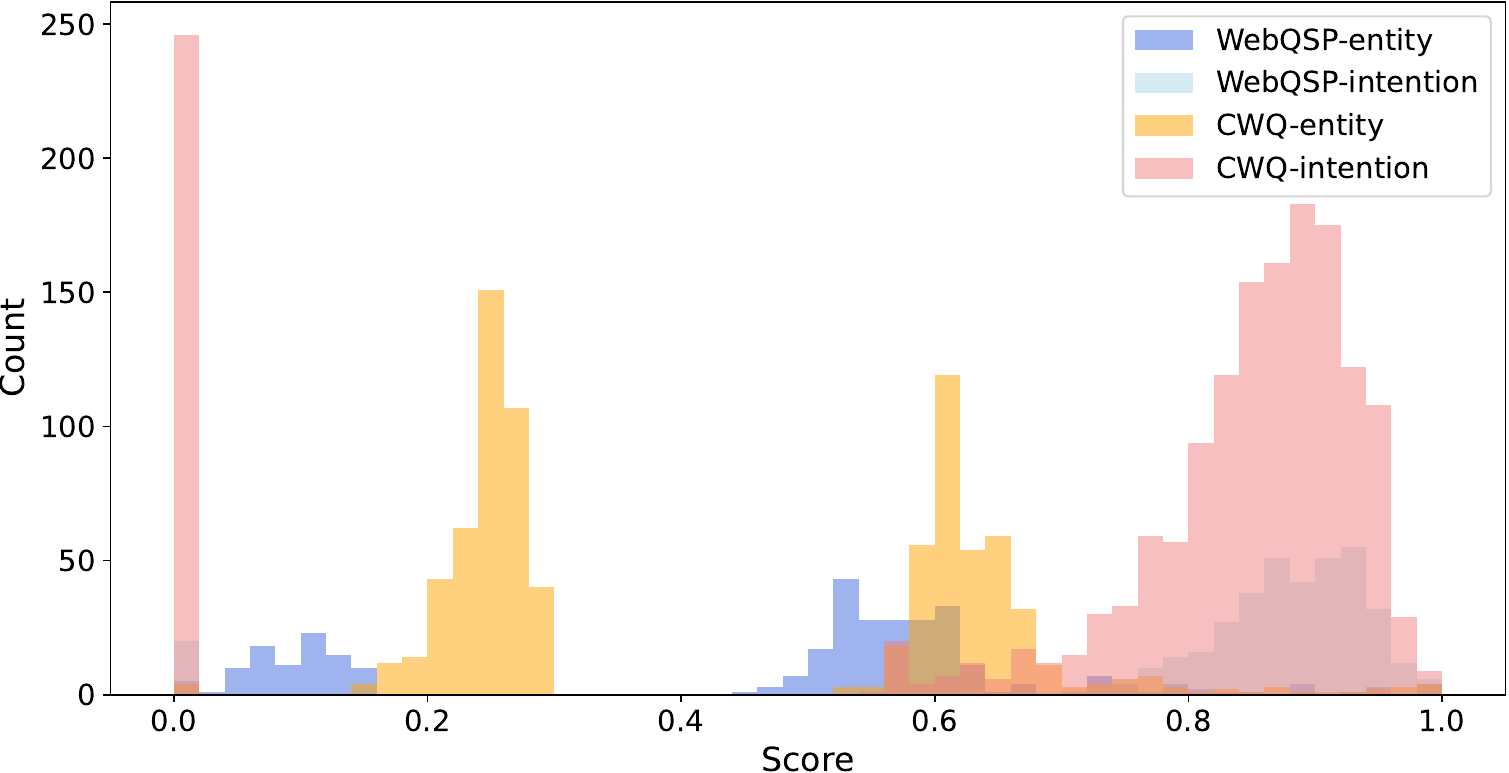}
    \caption{Distribution of entity score and intent score.}
    \label{fig:distribution_entity_intention}
\end{figure}

The distribution of entity scores and intent scores is illustrated in Figure \ref{fig:distribution_entity_intention}. Two key observations can be made:

1. The intent score distribution exhibits a polarized pattern. This is attributed to the presence of numerous semantically similar predicates in Freebase, which leads to higher scores.

2. The entity score distribution is comparatively less dispersed. Although entity name duplication is common, the questions in the dataset typically involve well-known entities, resulting in lower ambiguity scores.

\subsection{Ablation Study}

\begin{table}[ht]
    \caption{Ablation Study}
    \centering
    \scalebox{1}{
    \begin{tabular}{lcccc}
        \toprule
        \multicolumn{1}{c}{\multirow{2}{*}{\textbf{Method}}} & \multicolumn{2}{c}{\textbf{WebQSP}} & \multicolumn{2}{c}{\textbf{CWQ}} \\ \cline{2-5} 
        \multicolumn{1}{c}{} & \textbf{Overall} & \textbf{RHits@1} & \textbf{Overall} & \textbf{EM} \\ \midrule
        Ours w/ GPT-4o & 83.00 & 83.00 & 52.46 & 66.33 \\
        \quad w/o entity score & 75.27 & 75.27 & 48.93 & 69.55 \\
        \quad w/o intent score & 74.45 & 74.45 & 49.58 & 69.00 \\
        \quad w/o plugin & 73.15 & 73.77 & 46.88 & 64.00 \\ \midrule
        Ours w/ FT open-LLM & 73.77 & 73.77 & 55.86 & 62.33 \\
        \quad w/o entity score & 69.72 & 69.72 & 53.10 & 64.00 \\
        \quad w/o intent score & 72.47 & 72.47 & 54.36 & 67.00 \\
        \quad w/o plugin & 66.33 & 66.33 & 52.43 & 65.00 \\ \bottomrule
    \end{tabular}
    }
    \label{tab:ablation_study}
\end{table}

To investigate the impact of entity clarification and intent clarification functions in our plugin, we conducted ablation studies on both components. Additionally, we performed experiments with both functions disabled (w/o plugin), where only the \texttt{AskForClarification} tool was added to the prompt text, allowing the LLM to determine when to ask for clarification during interactions autonomously. The results are shown in Table \ref{tab:ablation_study}.

The results indicate that the plugin provides a more substantial performance boost for GPT-4o compared to the FT model, suggesting that GPT-4o has superior instruction-following capabilities. Furthermore, the entity clarification and intent clarification components demonstrate comparable importance in contributing to the overall performance.

\subsection{Error Analysis}

\begin{table}[ht]
    \caption{Error Analysis}
    \centering
    \scalebox{1}{

    \begin{tabular}{lcc}
        \toprule
        \textbf{Error Type}            & \textbf{GPT-4o}     & \textbf{FT open-LLM}    \\ \midrule
        Predicate Recognition & 25         & 19        \\
        Reasoning Errors      & 18         & 39        \\
        Constraint Missing    & 29         & 17        \\
        Golden Incorrect      & \multicolumn{2}{c}{10} \\
        Other                 & 18         & 25        \\ \bottomrule
    \end{tabular}

    }
    \label{tab:error_analysis}
\end{table}
To systematically assess our method's deficiencies, we first combine WebQSP and CWQ datasets (WCWQ) and randomly select 100 error instances where both GPT-4o and FT open-LLM failed to produce correct answers. The aggregated statistical findings are detailed in Table \ref{tab:error_analysis}.

\begin{itemize}[noitemsep, leftmargin=*]

\item \textbf{Predicate Recognition} refers to failures in identifying the correct predicates in the knowledge graph. For instance, failing to map ``locate in'' to the appropriate predicate ``organization.organization.headquarters''.

\item \textbf{Reasoning Errors} indicates insufficient semantic understanding of the KG schema by the LLM, such as misinterpreting predicate directions, resulting in incomplete or inaccurate SPARQL queries.

\item \textbf{Constraint Missing} denotes failures in properly handling constraints. For example, in the question ``who is the president of peru now?'', the temporal constraint ``now'' is not correctly processed in the query.

\item \textbf{Golden Incorrect} represents cases where the ground truth annotation is questionable. For instance, in ``what places make up new england?'', the golden SPARQL query constrains entity types to ``State'', which is not readily inferrable from the question.

\item \textbf{Other} encompasses errors that cannot be categorized under the aforementioned types.

\end{itemize}

\section{Conclusion}

CLEAR-KGQA introduces a clarification-enhanced ambiguity resolution framework for KBQA. It features a Bayesian rule-based plugin that quantifies entity and intent ambiguity, enabling the LLM to explicitly prompt users for clarification during multi-round interactions with a knowledge base. Furthermore, based on the interaction records, we generated an unambiguous KBQA dataset. Experimental results demonstrate that even baseline models can achieve promising performance on this dataset.

\section*{Acknowledgment}

This work was supported by the National Key Research and Development Program of China under Grant No. 2023YFC3304404.

\bibliographystyle{plain}
\bibliography{references}

% \appendix
% \include{chap/99-appendix}

\end{document}